\documentclass[runningheads]{llncs}
\usepackage{graphicx}
\usepackage{xcolor}
\usepackage{amsmath,amsfonts,amssymb,amsthm,mathtools}

\theoremstyle{definition}

\theoremstyle{definition}

\theoremstyle{remark}

\usepackage{hyperref}
\usepackage{cleveref}
\usepackage{booktabs}
\usepackage{graphicx} 
\usepackage{tabularx}
\usepackage{array}
\newcolumntype{Y}{>{\centering\arraybackslash}X}
\usepackage{array}
\usepackage{multirow}
\usepackage{algpseudocode}
\usepackage{algorithm} 
\algnewcommand{\algorithmicgoto}{\textbf{go to}}%
\algnewcommand{\Goto}[1]{\algorithmicgoto~\ref{#1}}%
\usepackage{enumitem}
\usepackage{listings}
\usepackage[frozencache,cachedir=.]{minted}

\newcommand{\MV}{\texttt{ModelVerification.jl} }
\usepackage{pifont}
\newcommand{\vmark}{\ding{51}}

\usepackage{graphicx} 
\usepackage{hyperref} 
\usepackage[firstpage]{draftwatermark} 

\SetWatermarkAngle{0}

\SetWatermarkText{\raisebox{12.5cm}{
\hspace{-0.12cm}
\href{https://zenodo.org/records/15319912}{\includegraphics{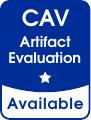}}
\hspace{8.5cm}
\includegraphics{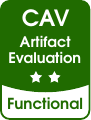}
}}

\makeatletter
\newcommand{\printfnsymbol}[1]{%
  \textsuperscript{\@fnsymbol{#1}}%
}
\makeatother
\begin{document}
\title{ModelVerification.jl: a Comprehensive Toolbox for Formally Verifying Deep Neural Networks}
\titlerunning{ModelVerification.jl}
%
\author{Tianhao Wei\inst{1} \and
Hanjiang Hu\inst{1*} \and
Luca Marzari\inst{1,2}\thanks{These authors contributed equally to the paper.} \and
Kai S. Yun\inst{1}\printfnsymbol{1} \and \\ Peizhi Niu\inst{1*} \and Xusheng Luo\inst{1} \and
Changliu Liu\inst{1}}

\authorrunning{T. Wei, H. Hu, L. Marzari, K. S. Yun,  P. Niu, X. Luo  and C.Liu} 

%

\institute{Carnegie Mellon University, Pittsburgh, USA\\
\email{\{twei2,hanjianh,xushengl,cliu6\}@andrew.cmu.edu} \and
University of Verona, Verona, Italy
\email{luca.marzari@univr.it}
}

\maketitle              

\begin{abstract}
Deep Neural Networks (DNN) are crucial in approximating nonlinear functions across diverse applications, ranging from image classification to control. Verifying specific input-output properties can be a highly challenging task due to the lack of a single, self-contained framework that allows a complete range of various model architecture and input-output properties. To this end, we present \texttt{ModelVerification.jl (MV.jl)}\footnote{\url{https://github.com/intelligent-control-lab/ModelVerification.jl}}, the first comprehensive, cutting-edge toolbox that contains a suite of state-of-the-art methods for verifying different types of DNNs and input-output specifications. This versatile toolbox is designed to empower developers and machine learning practitioners with robust tools for verifying and ensuring the trustworthiness of their DNN models.
\footnote{Accepted by CAV 2025.}

\end{abstract}
\section{Introduction}

The use of Deep Neural Networks (DNNs) is becoming increasingly prominent in several applications, including image classification \cite{image2,image1,li2018flow,han2020design}, autonomous navigation \cite{chen2015deepdriving,navigationCurriculum,navigation}, robotics \cite{Amir2023,manipulation1,manipulation2,wei2024metacontrol}, and control \cite{liu2023safe,wei2019safeControl,wei2022safeControl,xiang2018reachability}. The main characteristic of these functions is their ability to approximate complex nonlinear functions often employed in solving these tasks. Nonetheless, while these functions are very efficient, their opaque nature can result in unpredictable and potentially unsafe behavior when small changes in the input, often imperceptible to the human, are performed. Broadly speaking, these functions are subject to so-called ``adversarial inputs" \cite{adversarial} that can make them behave unsafely both for the system itself and especially for those around them. Hence, given the applications of DNNs in safety-critical contexts where human life is potentially at risk, the need to obtain formal guarantees about the safety of these systems arises and is of paramount importance.

To address this issue, the research field of Formal Verification (FV) of DNNs \cite{LiuSurvey}, has emerged as a valuable solution to provide formal assurances on the safety aspect of these functions before the actual deployment in real scenarios. The main goal of FV is to prove (or falsify) a desired input-output relationship (safety property) for a given DNN. More specifically, many methods have been developed to formally verify collision avoidance tasks with standard Feed Forward  Neural Networks (FFNNs) or robustness in image classification with Convolutional Neural Networks (CNNs) using reachability analysis \cite{Ai2,eran,NSVerify,ExactReach,Sherlock,imageStar}, optimization techniques \cite{ILP,MILP,Reluplex,Marabou}, or combining the two approaches \cite{VeriNet,Venus,crown,bcrown,GPCcrown}. Recently, there have also been techniques to find not only an individual violation point in the property's input domain but also to enumerate entire regions that may lead to unsafe behaviors to repair the network in those specific areas \cite{CountingProVe,eProve,yang2022neural}. 

Despite the considerable advancements made by FV over the years, given the NP-complete nature of the problem \cite{Reluplex}, there are still several remaining issues, such as scalability, that limit the application of these systems in very large and complex real-world scenarios. Moreover, another limitation of applying FV in realistic scenarios is that existing toolboxes tailored themselves to different assumptions of tasks or properties.
Hence, the complexity of the verification landscape in literature implies that users may need to switch between toolboxes or solvers when they intend to employ diverse verification approaches. This necessity poses a significant challenge, as such transitions are often neither convenient nor user-friendly. As a result, using an in-depth and comprehensive pre-deployment formal verification process is hard to achieve, and often, the only guarantees of safety rely on pure empirical evaluations.

\begin{figure}[h!]
    \centering
    \includegraphics[width=\linewidth]{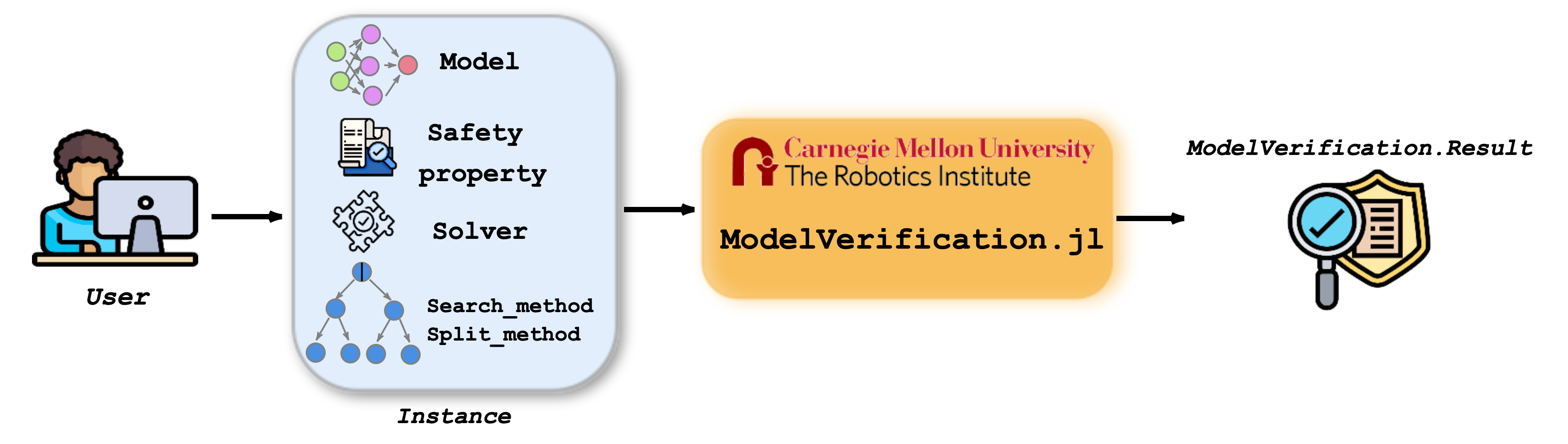}
    \caption{The user specifies the network, the safety property to check, and the  solver. \MV provides an assertion of whether the safety property holds.}
    \label{fig:overview}
\end{figure}

To this end, in this work, we present \MV (Fig.\;\ref{fig:overview}), the first comprehensive cutting-edge toolbox that contains a suite of state-of-the-art methods for verifying different types of DNNs and safety specifications.

\vspace{-3mm}
Our toolbox targets two distinct user categories within the formal verification domain. The first target audience comprises individuals who are relatively new or considered ``outsiders" to the FV world. For this latter, our toolbox is designed to be \textit{user-friendly}, accompanied by complete and comprehensive documentation of all the methods developed. Hence, we provide an accessible and educational resource for those looking to learn the intricacies of the field.
Concurrently, the second audience consists of researchers already well-versed in FV practices. Our toolbox offers valuable resources to even the sophisticated requirements of experienced practitioners. More specifically, our toolbox is written in Julia \cite{julia} language, ideal for specifying algorithms in human-readable form, and with the key \textit{``multiple dispatch"} feature, that enhances the development of an elegant and highly modularized design for \MV.  From this design, expert users can access the combination of great performing solvers on par with state-of-the-art ones (i.e., we focus not only on user-friendliness but also on toolbox efficiency) and even the possibility of implementing novel strategies of different natures (e.g., combining optimization and reachability) all in a single comprehensive toolbox.

\textit{\textbf{Yet another FV toolbox?}} The Formal Verification of DNNs is increasingly becoming essential for providing provable guarantees of deep learning models. We refer the interested reader to the following article for a complete taxonomy of the various state-of-the-art methods \cite{LiuSurvey}. In addition to these works mentioned in \cite{LiuSurvey}, it is important to mention recent methods such as Verinet \cite{VeriNet}, MN-BaB\cite{MN-BaB}, $\alpha$-$\beta$-CROWN \cite{bcrown,acrown,crown}, which provide the ability to test more complex properties, such as semantic perturbation, in addition to the classic methods based on regression and classification tasks. However, although $\alpha$-$\beta$-Crown, for instance, was the top performer in the last three years of the NN verification competition \cite{VNNComp3years}, it lacks support for novel types of DNNs such as Neural Ordinary Differential Equations (NeuralODEs) \cite{neuralODE}. To this end, a recent toolbox, NNV 2.0 \cite{NNV2.0}, has arisen to overcome this limitation. Still, the latter presents a lack of support for different deep learning models such as Residual Neural Networks (ResNets) \cite{ResNet}, confirming the non-existence of a single, self-contained framework that allows a complete range of verification types. We then have a range of toolboxes such as Juliareach \cite{Juliareach}, Sherlock \cite{Sherlock}, jax\_verify \cite{jax_verify}, ReachNN \cite{reachnn}, and RINO \cite{rino} that mainly focus on specific verification of DNNs (e.g., for Control Systems); and as also pointed out in \cite{NNV2.0}, either they present lack of support for different types of DNNs or are no longer maintained. In contrast, our toolbox covers major state-of-the-art verifiers, including $\alpha$,$\beta$-CROWN \cite{bcrown,acrown,crown}, Image-Star \cite{imageStar}, DeepZ, Zonotope \cite{Ai2}, and different layer types as mentioned before, enabling the user to pick the most appropriate solver for the given problem. Hence, \MV is the first self-contained toolbox that supports different verification and safety specification types designed to empower developers and machine learning practitioners with robust tools for verifying and ensuring the trustworthiness of their DNN models.
\section{Toolbox Features}\label{sec:tb_features}

To overcome the limitations presented in the previous section, we now discuss the main features and the improvement of \texttt{ModelVerification.jl} over the state-of-the-art in four macro categories:
\newline

\noindent \textbf{1) \textit{Comprehensiveness}.} As previously discussed, a notable constraint of using pre-deployment FV arises from the lack of a unified framework for verifying a broad spectrum of safety models and properties. Notably, existing solvers employ distinct representations for property verification, or they exclusively address particular categories of DNNs, thereby complicating the transition between tools. Consider a scenario where we have a collection of models encompassing both ResNets \cite{ResNet} and NeuralODEs \cite{neuralODE} alongside a set of safety properties to be verified. If, for instance, we opt to use the state-of-the-art $\alpha$-$\beta$-CROWN method \cite{acrown,bcrown,crown}, it exclusively supports the verification of the former type of networks. Meanwhile, for the latter, an alternative solver such as NNV 2.0 \cite{NNV2.0} becomes necessary. A critical constraint lies in the fact that these distinct solvers may be implemented following different design architecture strategies or even in different programming languages, as exemplified in this case, where the first solver is coded in Python while the second one is in Matlab. Consequently, accomplishing the verification process, in this case, entails the user's proficiency in both languages and a comprehensive understanding of how safety properties are encoded within the respective toolboxes.

To address such an issue, our primary objective is to provide the community with a tool of maximal comprehensiveness. We report in Tab. \ref{tab:features} the main features supported by \MV.

\begin{table}[h]
    \vspace{-4mm}
    \scriptsize
\begin{tabularx}{\linewidth}{lX}
        \textbf{Feature} & \textbf{\MV support}   \\
        \toprule
        Neural Network & FFNN, CNN, ResNet, and NeuralODE\\
        \midrule
        Activation functions & ReLU, Sigmoid, Tanh\\
        \midrule
        Layers type & FC, Linear, ReLU, MaxPool, AvgPool, Conv, Identity, BatchNorm, Skip, Parallel\\
        \midrule
        Geometries Representation & Hyperrectangle, Polytope, Zonotope, Star, ImageStar, ImageZono, Image Convex Hull, Taylor Model Reachable Set\\
        \midrule
         Verification &  Safety, Robustness, Adversarial attack, VNNLIB, Enumeration of (un)safe regions\\
        \midrule
        Reachable set visualization &  Layer-by-layer, Exact and Over-approximation visualization\\
        \bottomrule
    \end{tabularx}
    \vspace{2mm}
    \caption{Features supported by \MV.}
    \label{tab:features}
    \vspace{-8mm}
\end{table}
Hence, the main purpose of our toolbox is to provide the possibility to verify all different types of neural networks, starting from the classical FFNNs and CNNs up to the more complicated ResNets and NeuralODEs. Also of primary importance is the support of general squashing activation functions, such as Tanh and Sigmoid, in addition to the standard ReLU. Moreover, we decide to write \MV in Julia for the following reasons: 
\begin{itemize}
    \item Julia is a language specifically designed for scientific computation, which combines the efficiency of C and the flexibility of Python.
    \item We have an ample range of libraries available for operations with various complex geometric figures (e.g., \textit{LazySets} \cite{Lazysets}). While this gives us prominent performance, it also allows us to encode a wide range of safety properties with consistent geometric representations, resulting in a unified framework as desired.
    \item Julia's \textit{``multiple dispatch"} feature allows us to adopt a uniform abstract pipeline such that different solvers can share the same function interface. The pipeline is both efficient and easy to follow.
\end{itemize} 

\begin{figure}[h!]
    \centering
    \includegraphics[width=0.85\textwidth]{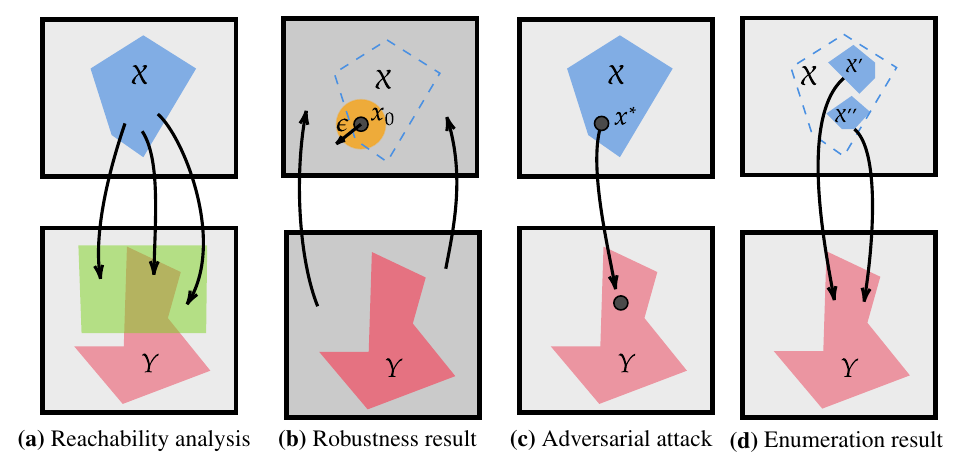}
    \vspace{-3mm}
    \caption{Different types of verification supported in \MV. $\mathcal{X}$ represents the safety property's domain, while $\mathcal{Y}$ the undesired reachable set.}
    \label{fig:verification_types}
\end{figure}

\begin{figure}[h!]
    \vspace{-3mm}
    \centering
    \includegraphics[width=\textwidth]{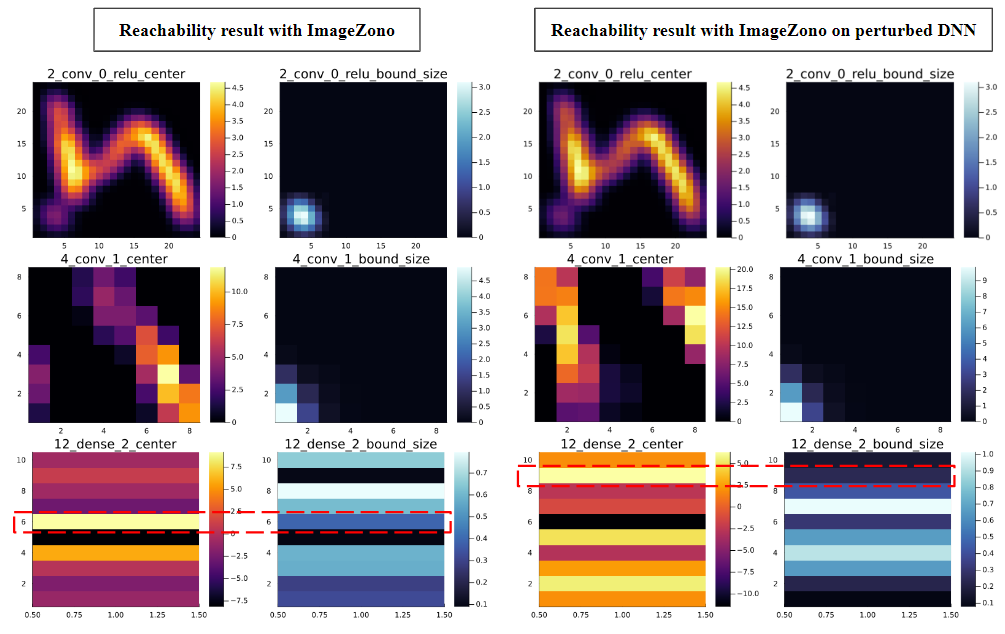}
    \caption{Explanatory example of visualization of the reachable set layer-by-layer using \MV for a specific robustness verification instance of MNIST dataset. In this example, a single image representing the ``five" handwritten digit and a local perturbation in the bottom left corner of the figure is considered. On the left part of the image, we report layers 2, 4, and 12's reachable sets computed using ImageZono, where each reachable set is visualized using its center and the bound size using a heatmap. On the right, the reachable sets computed using ImageZono for a perturbed DNN are visualized. A convolutional layer and the last dense layers of the DNN are perturbed to visualize their effect on the final prediction. In the last row, we highlighted the predicted class in red. Crucially, we can notice a large scale in correspondence to the lighter row, meaning that the noise is larger.}
    \label{fig:visualization}
\end{figure}

Addressing the comprehensiveness, our toolbox provides the ability to perform several types of verification (Fig. \ref{fig:verification_types}), not only safety and robustness, using reachability analysis (Fig. \ref{fig:verification_types}a-b), but also the possibility to perform adversarial attack (Fig. \ref{fig:verification_types}c) --by exploiting one of the main methods, such as Fast Gradient Sign Method (FGSM), Projected Gradient Descent (PGD) attack, and Auto-PGD. Moreover, our toolbox includes recent exact and approximation methods \cite{yang2022neural,CountingProVe,eProve} even to enumerate the set of (un)safe regions of a given safety property (Fig. \ref{fig:verification_types}d). Finally, \MV provides the user with visual representations of the intermediate results of the verification process (i.e., the reachable sets) as depicted in Fig. \ref{fig:visualization}. 
We also introduce a new type of input set, ImageConvexHull, which contains all possible interpolations of the given seed images. ImageConvexHull is particularly useful for semantic perturbations such as occlusion, rain, fog, and shadow.

All of these features enable \MV to verify different types of networks and properties in a single framework. We report in Tab. \ref{tab:improvement} the improvements of our toolbox with respect to  $\alpha$-$\beta$-CROWN \cite{acrown,bcrown,crown}, NNV 2.0 \cite{NNV2.0}, and MN-BaB \cite{MN-BaB} methods, considered state-of-the-art for formal verification of neural networks.

\begin{table}[h]
    \centering
    \scriptsize
    \begin{tabularx}{\linewidth}{@{}>{\centering\arraybackslash \hsize=.5\hsize}X>{\centering\arraybackslash \hsize=1.25\hsize}X>
    {\centering\arraybackslash \hsize=1.25\hsize}X>
    {\centering\arraybackslash \hsize=1.25\hsize}X>
    {\centering\arraybackslash \hsize=1.25\hsize}X>
    {\centering\arraybackslash \hsize=1.25\hsize}X@{}
    }
        
        \textbf{Features}& &  &\textbf{Toolbox - Solver}  &  \\ 
        \toprule
         & $\alpha$-$\beta$-CROWN & NNV 2.0 & MN-BaB  & \texttt{MV.jl}   \\
        \toprule
        Standard Layers & \vmark & \vmark & \vmark & \vmark \\
         \midrule
        General Comp. Graph & \vmark & \vmark & \vmark & \vmark \\
         \midrule
        General non-linearities & \vmark & \vmark & \vmark & \vmark \\
        \midrule
        GPU support & \vmark &   & \vmark & \vmark \\
        \midrule
        Reachable set vis. &  & \vmark &  & \vmark \\
         \midrule
        Input sets & $L_p$-ball, VNNLIB format & $L_\infty$-ball, Zonotope, Star, Polyhedron, VNNLIB format & $L_\infty$-ball, Zonotope, VNNLIB format & $L_p$-ball, Polytope, Zonotope, Star, ImageConvexHull, VNNLIB format \\
        \midrule
        Solvers & $\alpha$-$\beta$-CROWN, IBP, CROWN, MIP & Zonotope, Star, NeuralODE & IBP, Zonotope, MN-BaB & $\alpha$-$\beta$-CROWN, IBP, CROWN, Zonotope, Star, MN-BaB, NeuralODE \\
        \bottomrule
    \end{tabularx}
    \vspace{2mm}
    \caption{Comparison between \MV and existing state-of-the-art toolboxes.}
    \label{tab:improvement}
    \vspace{-8mm}
\end{table}

\vspace{2mm}
\noindent \textbf{2) \textit{User-friendliness}.} Another major aspect of our toolbox is the ease of use. Our toolbox only requires several lines of code to formulate a verification problem in most cases. We provide comprehensive documentation, facilitating the use of the toolbox through detailed explanations and tutorials and, for Python users, a compiled library of this package such that they can directly call the package from Python itself.

To provide the reader with an intuition of the user-friendliness of our toolbox, let us consider a verification task that considers verifying a ResNet-based NeuralODE. 
Due to the modularized design chosen for \MV and the possibility of combining different solver strategies, we obtain a toolbox that encapsulates a vast range of verification scenarios, avoiding any model architecture modification potentially required in other solvers to meet their specific design. 
\begin{minted}[frame=lines,fontsize=\tiny]{julia}
using ModelVerification as MV
model = MV.get_resnet_model("path_to_model")
input_set  = Hyperrectangle(low=[0.9, -0.1], high=[1.1, 0.1])
output_safe_set = Hyperrectangle(low = [2.2, 2.2], high = [2.8, 3.2])
search_method = BFS(max_iter=10, batch_size=1)
split_method = Bisect(1)
prop_method = ODETaylor(t_span=1.0)
verify(search_method, split_method, prop_method, ODEProblem(model, input_set, output_safe_set))
\end{minted}

More specifically, in \MV with a few simple lines of code, reported in the listing above, we can load the desired model and perform the required type of verification, regardless of the dataset we want to use. In particular, our toolbox provides a set of model converters commonly used in the literature, such as ONNX, TensorFlow (Keras), and PyTorch, to name a few, to Flux models, a Julia library for machine learning that contains an intuitive way to define models, just like mathematical notation. In addition, Flux allows differentiable programming of cutting-edge models such as neuralODEs, typically not supported by state-of-the-art (e.g., $\alpha$-$\beta$-CROWN) methods, as previously discussed. As we can notice in the code above, the high-level language exploited in Julia allows for an easy understanding of what is being performed in the verification phase. Moreover, to increase even further the level of clarity, \MV provides the possibility to obtain a set of intra-layer representations of reachable sets obtained during the verification process, as shown in Fig. \ref{fig:visualization}. This visualization shows how the perturbations ``diffuse" during the reachability analysis, and how does it affects the final prediction, providing a human conceivable robustness.

\begin{figure}[t]
    \centering
    \includegraphics[width=0.95\textwidth]{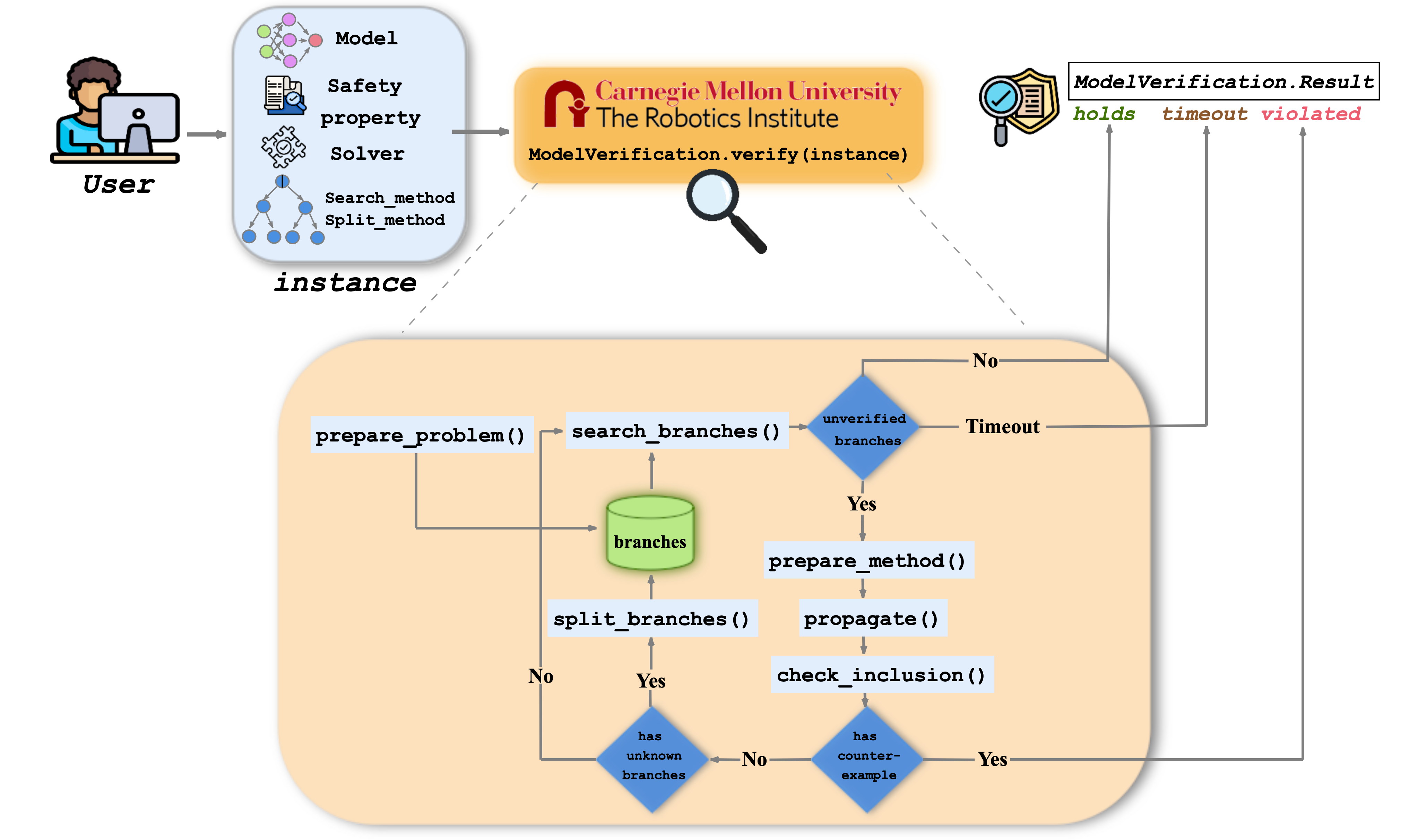}
    \caption{Computational flow of \MV. The user provides the verification problem, including the model, the input set, and the desired output property. Our toolbox follows a branch and bound scheme to divide and conquer the problem. A result will be returned to verify or falsify the property if not timed out.}
    \label{fig:flow}
\end{figure}

\noindent \textbf{3) \textit{Extensibility}.} Our toolbox follows a highly modularized design, making it easy to understand and customize. Specifically, based on the \textit{``multiple dispatch"} feature previously mentioned, we developed straightforward and easy-to-follow implementations. We abstract out a general pipeline and modularize \texttt{MV.jl} following the standard Branch-and-Bound (BaB)~\cite{liu2021algorithms} paradigm. In detail, all the verification algorithms implemented in our toolbox divide the hard-to-verify problem into easier problems and proceed to verify the single easier subparts. This results in the possibility of choosing and combining different existing solvers provided in the toolbox to solve each part of the verification process optimally. 

In the literature, it is worth noting that some solvers work best exploiting GPU computation, while others heavily rely on the CPU. Crucially, our toolbox supports both GPU and CPU-based methods. This dual support, in combination with an elegant, highly modularized, and well-documented BaB design, enables a key feature of our toolbox with respect to other state-of-the-art methods, as such, the possibility to combine different solvers for any verification purpose. 
We report in Fig. \ref{fig:flow} a high-level overview of the computational flow of \texttt{ModelVerification.jl}. As discussed, each base submethod that composes the \textit{verify} function is highly customizable based on the user's necessity. Based on \texttt{MV.jl},  neural control barrier functions \cite{hu2024verification} and neural Hamilton-Jacobi Reachability  value functions \cite{yang2024scalable} can be verified. \\
\newline
\noindent \textbf{4) \textit{Efficiency}.} Besides prioritizing user-friendliness, the last main feature of \MV is concerned with efficiency. Our toolbox provides significant improvements over the first Julia toolbox ever for FV of DNNs  called NeuralVerification.jl (NV.jl) \cite{LiuSurvey}. In detail, NV.jl is written in Julia to provide the community with pedagogical and immediate-to-understand implementations, similar to our goal. However, given the pedagogical nature adopted, performance is suboptimal. In contrast, based on the architectural design choices for our toolbox, we are able to achieve user-friendly implementations and, at the same time, efficient results --in terms of verification time and scalability-- comparable to, or in certain cases, surpassing those achieved by state-of-the-art solvers, as shown in Sec. \ref{sec:evaluation}. Recently,~\cite{luo2024certifying} efficiently verifies semantic perturbations on images using zonotope-based reachability analysis, while~\cite{cheng2024robust} verifies robust model predictive control leveraging the efficient CROWN~\cite{crown} implementation in \texttt{ModelVerification.jl}.

\section{Evaluation}\label{sec:evaluation}

This section demonstrates how versatile \MV is in encoding various input-output specifications for various tasks, as well as testing the performance of our toolbox in standard benchmarks from the VNN competition \cite{VNNComp3years} to showcase the efficiency.

\subsection{Empirical evaluation on VNN benchmarks}
\begin{figure}[h!]
    \centering
    \includegraphics[width=0.8\linewidth]{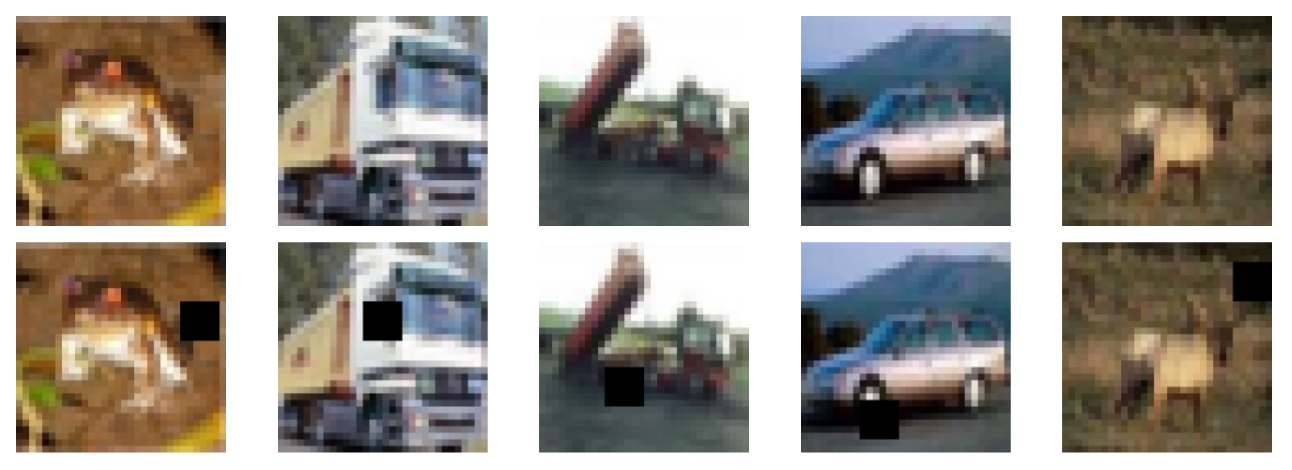}
    \caption{Examples of the original and the occluded images used for the ResNet verification process.}
    \label{fig:examples}
    \vspace{-3mm}
\end{figure}

 The first part of our evaluation concerns the robustness of trained ResNets, in particular, ResNet2b and ResNet4b. This verification is a valuable benchmark of scalability, particularly difficult to verify due to the large number of parameters contained in these architectures.
 In detail, ResNet2b comprises two residual blocks composed of five convolutional layers plus two linear layers, while ResNet4b has four residual blocks with nine convolutional layers and two linear layers. For this evaluation, we use images from the CIFAR-10 dataset \cite{CIFAR} (composed by image size $32 \times 32$) with different occlusion perturbations. The occlusion adopted is a $6 \times 6$ black block and is randomly placed on the original image. We report in Fig. \ref{fig:examples} a set of explanatory images of the type of robustness tested.
In this study, a total of 100 images were subjected to verification using \textit{ImageZono} as the solver, and the outcomes are presented in Tab. \ref{tab:results}. All instances yielded deterministic results. Notably, the ResNet4b model, characterized by a greater number of layers and enhanced robustness, exhibited a higher number of \textit{holds} instances and thus a longer verification time compared to ResNet2b.

\begin{table}[t]
    \scriptsize
    \centering
    \begin{tabular}{cccccc}
        \toprule
         & Holds instance & Violated instance & Unknown instance & \#Parameters & Time \\
         \midrule
         ResNet2b & 54 & 46 & 0 & 112K & 93.51s \\
         ResNet4b & 72  & 28 & 0 & 123K & 1086.40s \\
         \bottomrule
    \end{tabular}
    \vspace{2mm}
    \caption{Verifying ResNet with occlusion perturbation.}
    \label{tab:results}
    \vspace{-3mm}
\end{table}

\begin{figure}
    \centering
    \includegraphics[width=0.9\linewidth]{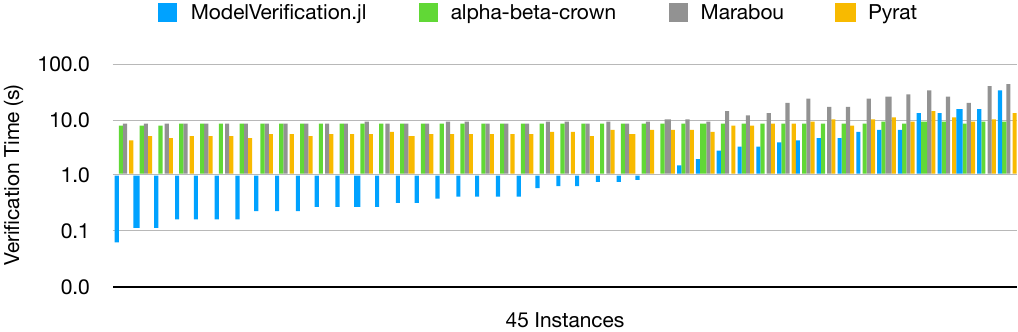}
    \caption{Verification time of 45 instances in ACAS Xu $\phi_1$. \MV is the fastest for most of the instances. The average verification time is \MV: 3.34s, \texttt{$\alpha$-$\beta$-CROWN}: 8.37s, \texttt{Marabou}: 13.60s, and \texttt{PyRat}: 9.12s.}
    \label{fig:acas}
\end{figure}


We then evaluate \MV on a subset of benchmarks from VNN-COMP’23 \cite{vnnComp}, ACAS Xu property $\phi_1$ for $45$ different networks that have 13K parameters. We compare with the toolboxes that won the first three places: \texttt{$\alpha$-$\beta$-CROWN}, \texttt{Marabou}, and \texttt{PyRat}. We run our toolbox on an AWS m5 instance following the same VNN-COMP setup~\cite{vnnComp} as other toolboxes. The results of other toolboxes are directly from the competition. Detailed setting can be found in our toolbox repository. As shown in Fig.~\ref{fig:acas}, our toolbox is faster than other toolboxes for most instances of this property, showcasing its efficiency. 



\section{Discussion}\label{sec:discussion}

We introduced \texttt{ModelVerification.jl}, a comprehensive toolbox for verifying deep learning models. Our tool is the first cutting-edge toolbox containing a suite of state-of-the-art methods for
verifying DNNs, including the verification of feedforward, convolutional, ResNet \cite{ResNet}, and NeuralODEs. We believe the easy-to-follow implementation, combined with detailed documentation, provides a valuable and unique resource for using formal verification, even for people new to the subject. Moreover,  the wide range of geometries that can be employed to describe both safety properties and different types of verification problems allows even the most experienced users to be able to take full advantage of this tool.
For the future development of this toolbox, we want to further optimize the performance of the implemented solvers, including making the code more GPU-friendly, optimizing the general structure to reduce redundant computation, supporting more branching algorithms and solvers, and optimizing memory cost as well as performing a more comprehensive comparison with other state-of-the-art toolboxes.

\clearpage
\subsubsection*{Acknowledgements.} This work is in part supported by Boeing and in part supported by mobility grants for non-EU destinations of the University of Verona's Doctoral School. Tianhao leads the design, implementation and evaluation of the toolbox. Hanjiang, Peizhi, and Xusheng contribute to the toolbox implementation and evaluation. Luca and Tianhao lead the paper writing. Kai and Luca lead the documentation and tutorial. 
%
%
%
\bibliographystyle{splncs04}
\bibliography{mybibliography}

\end{document}